\documentclass[10pt]{article} % For LaTeX2e
\usepackage[preprint]{tmlr}
\usepackage{amsmath,amsfonts,bm}

\def\eqref#1{equation~\ref{#1}}
\def\1{\bm{1}}

\DeclareMathAlphabet{\mathsfit}{\encodingdefault}{\sfdefault}{m}{sl}
\SetMathAlphabet{\mathsfit}{bold}{\encodingdefault}{\sfdefault}{bx}{n}

\usepackage{url}
\usepackage{hyperref}
\usepackage{graphicx}
\usepackage{algorithm}
\usepackage{algpseudocode}
\usepackage{caption}

\title{Transferring Multiple Policies to Hotstart Reinforcement Learning in an Air Compressor Management Problem}

\author{\name H\'el\`ene Plisnier \email helene.plisnier@vub.be \\
      \addr Vrije Universiteit Brussel
      \AND
      \name Denis Steckelmacher \email denis.steckelmacher@vub.be \\
      \addr Vrije Universiteit Brussel
      \AND
      \name Jeroen Willems \email jeroen.willems@flandersmake.be \\
      \addr FlandersMake
      \AND
      \name Bruno Depraetere \email bruno.depraetere@flandersmake.be \\
      \addr FlandersMake
      \AND
      \name Ann Now\'e \email ann.nowe@vub.be\\
      \addr Vrije Universiteit Brussel}

\def\month{MM}  % Insert correct month for camera-ready version
\def\year{YYYY} % Insert correct year for camera-ready version
\def\openreview{\url{https://openreview.net/forum?id=XXXX}} % Insert correct link to OpenReview for camera-ready version

\begin{document}

\maketitle

\begin{abstract}
Many instances of similar or almost-identical industrial machines or tools are often deployed at once, or in quick succession. For instance, a particular model of air compressor may be installed at hundreds of customers. Because these tools perform distinct but highly similar tasks, it is interesting to be able to quickly produce a high-quality controller for machine $N+1$ given the controllers already produced for machines $1..N$. This is even more important when the controllers are learned through Reinforcement Learning, as training takes time, energy and other resources. In this paper, we apply Policy Intersection, a Policy Shaping method, to help a Reinforcement Learning agent learn to solve a new variant of a compressors control problem faster, by transferring knowledge from several previously learned controllers. We show that our approach outperforms loading an old controller, and significantly improves performance in the long run.
\end{abstract}

\section{Introduction}
%\IEEEPARstart{T}{his}

\begin{figure}[!t]
\centering
\includegraphics[width=2.5in]{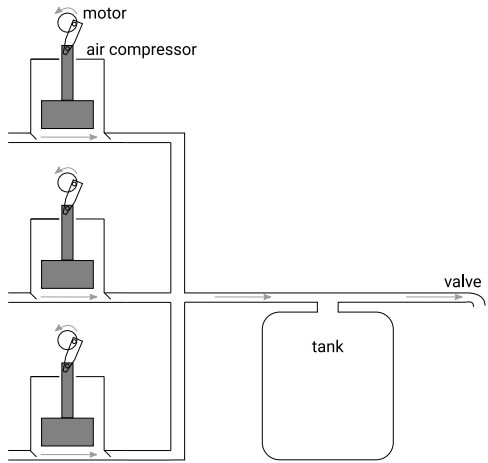}
% where an .eps filename suffix will be assumed under latex,
% and a .pdf suffix will be assumed for pdflatex; or what has been declared
% via \DeclareGraphicsExtensions.
\caption{Description of the setup. Three piston air compressors are filling a tank of varying size. The valve progressively opens or closes to simulate a demand.}
\label{fig_setup}
\end{figure}

A Reinforcement Learning algorithm takes an input, predicts an output, and then receives a scalar grading, or \emph{reward}, indicating how well it did; the objective of the Reinforcement learner is to maximize the rewards it gets. Reinforcement Learning (RL) is attractive when the solution to a problem is unknown or too difficult to implement by hand, but that it is possible to determine whether a solution is optimal or not. It has already been proven useful in a wide range of applications, such as industrial and healthcare problems: order dispatching \cite{Kuhnle2019}, datacenter cooling \cite{Lazic2018}, personalized pharmacological anemia management \cite{Gaweda2005}, etc. In this paper, we apply Soft Actor-Critic \cite{Haarnoja2018} (SAC), a well-known RL algorithm, to learn a controller in an air compressor management problem, which setup is detailed in Figure \ref{fig_setup}. Three piston compressors are expelling air in a tank; air is going out through a valve which progressively opens or closes following a demand curve. The compressors must provide a satisfactory pressure in between 3 and 5 bars, while conserving energy. The SAC Reinforcement Learning agent learns to control the three motors, each activating one of the piston compressor, while observing the current speed of the motors and the current pressure; the demand curve is unknown to the agent.

In addition, we consider the following problem: given that several controllers have already been learned, each in a different variant of the air compressor management problem, can a new controller be learned faster by exploiting already learned controllers? Variants of the air compressor management problem are generated by varying the volume of the tank, and the compressing power of the three compressors. To intelligently reuse the knowledge acquired by the previously learned controllers, Transfer Reinforcement Learning techniques can be used.

Transfer Learning \cite{Taylor2009} has the potential to make RL agents much faster at mastering new tasks, by allowing the reuse of knowledge acquired in previous tasks. A Transfer Learning setting often involves a \textit{source} task and a \textit{target} task; first, the agent learns the source task, then the knowledge acquired in the source task is intelligently leveraged by the agent while tackling the target task, using a Transfer Learning algorithm \cite{Taylor2009,Zhu2020}. Transferred knowledge can be the learned controller of the agent \cite{Taylor2007a,Fernandez2006,Brys2016}; learned skills \cite{Andre2002,Ravindran2003,Konidaris2007}, some general enough skill (like walking) to fit a large set of target tasks \cite{Tang2017}; parts of a modular neural network in which each module deals with a different aspect of the task \cite{Devin2017,Mirowski2018}. Methods to effectively transfer knowledge include reward shaping \cite{Brys2016}, controller reuse methods, which shape the agent's exploration strategy \cite{Fernandez2006,Griffith2013}, initializing a controller \cite{Taylor2007a} using information from the source task, and initializing parts of the target network with the source network \cite{Devin2017,Mirowski2018,Chaplot2016}, to only cite a few.

The Transfer Learning methods we evaluate in this paper is Policy Intersection (which we review in Section \ref{subsec:background_ps}), a Policy Shaping method using the previously learned controller to guide the exploration of the agent while it learns the new controller \cite{Garcia2015}. We apply Policy Intersection to effectively distill multiple transferred controllers in the learning process of a SAC agent on a new variant of the air compressor management problem. We empirically show that Policy Intersection outperforms simply loading a transferred controller in the new task, be it with one or multiple transferred controllers, and that leveraging multiple controllers performs significantly better than leveraging only one transferred controller.

\section{Background}
In this section, we introduce the Reinforcement Learning framework, the Soft Actor-Critic algorithm and Policy Intersection,  which we use in our experiments.
\subsection{Reinforcement Learning}
\label{sec:background_rl}

Reinforcement Learning techniques learn a controller that observes states of an environment, executes actions and receives rewards from the environment. An environment can be formulated using a Markov Decision Process (MDP) defined by the tuple $\langle S, A, R, T \rangle$ \cite{Bellman1957}, with the time discretized in \textit{time-steps}. At each time-step $t$, the agent takes an action $a_t \in A$ based on the current state $s_t \in S$, receives a new state from the environment $s_{t+1}  \in S$ and a scalar reward $r_t$, returned by a reward function $R(s_t, a_t, s_{t+1}) \in \mathcal{R}$.  For each state transition, a transition function $T(s_{t+1} | s_t, a_t) \in [0, 1]$ taking as input a state-action pair $(s_t, a_t)$ returns a probability distribution over new states $s_{t+1}$; $T$ is generally unknown to the agent. Both the actions set $A$ and states set $S$ can be finite or infinite.

 The amount of rewards received in the long run is formally defined as the discounted return $\mathcal{R}_t = \sum_t \gamma^t r_t$, where $\gamma \in [0, 1]$ is called the discount factor. The closer $\gamma$ is to 1, the more the agent will prefer long-term rewards to short-terms ones. At acting time, the agent select an action based on its controller, or policy $\pi$, which maps a state to a probability distribution over actions, with $\pi(a_t | s_t) \in [0, 1]$ denoting the probability of taking action $a_t$ given state $s_t$. After having executed action $a_t$ in state $s_t$, the environment returns a reward $r_t$, and the newt state $s_{t+1}$; the agent stores the data sample, or \emph{experience tuple} $(s_t, a_t, r_t, s_{t+1})$ in a buffer. This transaction between the agent and the environment, to go from state $s_t$ to $s_{t+1}$, occurs during one \emph{time-step}. At learning time, the agent updates its policy based on the experience tuples collected during acting time. The objective of an RL algorithm is to progressively modify $\pi$ until the optimal policy $\pi^*$ is found, which maximizes $E_{\pi^*}[\mathcal{R}_t]$, the expected discounted return \cite[ch. 3]{Sutton2000}.

We focus on episodic tasks, in which time is broken down into \textit{episodes}; in the problem we tackle in this paper, an episode ends and a new one starts once a fixed amount of time-steps have passed. At the beginning of each new episode, the agent is put back in an initial state, and must re-perform the task from the beginning. The performance of an RL agent, and ultimately of the algorithm it executes, is evaluated mainly through looking at its \textit{learning curve}. A learning curve shows, for each episode, the sum of all rewards collected during the episode. The higher the curve goes, the better the policy learned is. In addition, an experiment often consists in several \emph{runs} of an RL algorithm learning a task from beginning to end, for a number of episodes.

\subsection{Soft Actor-Critic}
\label{subsec:background_sac}
Soft Actor-Critic (SAC) is an off-policy actor-critic Reinforcement Learning algorithm, originally tailored for environments with continuous actions \cite{Haarnoja2018}. A way to categorize an RL algorithm is by whether it consists in an on-policy or off-policy method: on-policy algorithms update the policy from which actions are executed in the environment, generating data samples; off-policy algorithms update another policy, different from the one generating the samples \cite{Sutton2018}. In contrast to on-policy algorithms, off-policy algorithms tend to be more efficient at pursuing the optimal policy $\pi^*$, and are less prone to fall in local optima. Furthermore, Reinforcement Learning algorithms either learn a $Q$ (\emph{quality}) function, or a critic, from which a controller is indirectly derived, a policy $\pi$, or actor, which directly acts as a controller, or both. SAC is of the latter category.

SAC is based on the principle of maximum entropy: the agent maximises $E_{\pi}[\mathcal{R}_t + \alpha \mathcal{H}(\pi)]$, hence not only the expected discounted return, but also the entropy of its policy. This approach results in a policy acting as randomly as possible all throughout learning, which allows the agent to intensively explore its environment. Continuous actions are challenging for Reinforcement Learning, and not every algorithm is compatible with them. For instance, almost all RL algorithms for continuous actions need an explicit actor: there are very few critic-only algorithms for continuous actions \cite{Engel2005,Antos2007}. In SAC, $\pi$ is implemented as a Gaussian distribution, parameterized by a mean $\mu$ and a standard deviation $\sigma$, these two parameters being output by a neural network given a state $s_t$. At each time-step, an action $a_t$ is sampled from the Gaussian policy with mean $\mu$ and standard deviation $\sigma$.

In the next section, we review how a learned policy $\pi$ can be directly influenced by an external advisory policy to guide its exploration.

%In addition to a $Q$ function parameterized by $\theta$ and a policy $\pi$ parameterized by $\psi$, SAC also updates a state value function $V$ parameterized by $\phi$; all three functions are represented by neural networks. The $V$ network is trained by minimizing the error:

\subsection{Policy Shaping}
\label{subsec:background_ps}

 Policy Shaping lets an external advisory policy $\pi_A$ alter or determine the policy of the agent at acting time, either sporadically or continuously throughout learning. This general approach, shared with the Safe RL and the learning from human interactions RL subfields \cite{Garcia2015,Alshiekh2018,Griffith2013}, can easily be used for Transfer Learning purposes. In this section, we review Policy Intersection \cite[originally called ``Policy Shaping"]{Griffith2013,Cederborg2015}, a method to shape an RL agent's policy. The first work we know of that introduced the term ``Policy Shaping'' is \cite{Griffith2013}. However, the method presented (which we detail below) is very specific, while the term Policy Shaping could apply to a wide variety of methods \cite{Macglashan2017,Harutyunyan2014}. Since \cite{Griffith2013} introduces an element-wise multiplication between discrete probability distributions, and that this operation represents taking the intersection between the distributions, we rebaptized Griffith's formula Policy Intersection in this paper for clarity. Similarly, a Policy Union algorithm can be achieved by summing the probability distributions, although the qualification and evaluation of this approach is outside the scope of this paper.

 \subsubsection{Policy Intersection}

This second approach at Policy Shaping multiplies the $\pi_L$ and $\pi_A$ vectors together then normalizes the resulting policy vector at each timestep:

\begin{equation}
\label{eq:ps}
    a_t \sim \pi_L \times \pi_A = \frac{
        \overbrace{\pi_L(s_t) \, \pi_A(s_t)}^
        {\text{element-wise product}}
    } {
        \underbrace{\pi_L(s_t) \cdot \pi_A(s_t)}_
        {\sum_{a \in A} \pi_L(a | s_t) \pi_A(a | s_t)}
    }
\end{equation}

\noindent where $\pi_L(s_t) \cdot \, \pi_A(s_t)$ is the dot product of the two policies. At acting time, the agent samples an action $a_t$ from the mixture $\pi_L \times \pi_A$ of the agent's current learned policy $\pi_L(s_t)$ and the external advisory policy $\pi_A(s_t)$, instead of sampling only from $\pi_L(s_t)$. This product of probability distribution vectors amounts to taking the intersection between what the current agent's policy wants to do, and what the transferred policy would do in that state. As a result, although this formula was first introduced by \cite{Griffith2013} as ``Policy Shaping", we re-name it Policy Intersection in this paper for clarity. Policy Intersection allows $\pi_L$ and $\pi_A$ to more cooperatively select actions, with $\pi_A$ able to increase or decrease the probability of actions, but without fully determining the action. However, Equation \ref{eq:ps} can only be applied to tasks for which there is a finite set of actions, excluding environments with a continuous action space.

The Actor-Advisor \cite{Plisnier2019} is the first attempt made at using the Policy Intersection formula in Equation \ref{eq:ps} to achieve Transfer Learning. Their main contribution is to directly influence a Policy Gradient agent's policy with some off-policy external advice $\pi_A$ without convergence issue. However, that approach is also limited to discrete actions.

\section{Related Work}
\label{sec:sota}

Transferring knowledge in Reinforcement Learning potentially improves sample-efficiency, as it allows an agent to exploit relevant past knowledge while tackling a new task, instead of learning the new task from scratch \citep{Taylor2009}. Usually, we consider that the valuable knowledge to be transferred in Reinforcement Learning is the actual output of a reinforcement learner: a $Q$ function or a policy $\pi$ \cite[p.34]{Brys2016}. Some work also consider reusing skills, or \emph{options} \citep{Sutton1999}, as a transfer of knowledge across tasks \citep{Andre2002,Ravindran2003,Konidaris2007}. We focus on Transfer Learning methods directly reusing the policy learned in the source task. In this section, we sort previous work in categories related to the \emph{way} the source policy is transferred into the agent, and look at what is allowed to be different between the source task and the target task. The two predominant ways in which knowledge can be transferred are i) the source policy serves as a guide during exploration, ii) the source policy is used to train or initialize the agent, so that the agent actively imitates it.

\subsection{Learning}

Some techniques have proposed to let the source policy ``teach'' the agent to perform the target task, either by dynamic teaching, or by straightforward initialization. Imitation learning aims at allowing a student agent to learn the policy of a demonstrator, out of data that it has generated \cite{Hussein2017}. Similarly, policy distillation \cite{Bucila2006} can be applied to RL to train a fresh agent with one or several expert policies, hence resulting in one, smaller, potentially multi-task RL agent \cite{Rusu2015}.

Imitation learning and policy distillation are somewhat related to transfer learning \citep[p.24]{Hussein2017}, although imitation and distillation assume that the source and target tasks are the same, while transfer does not. The Actor-Mimic \citep{Parisotto2015} uses several DQN policies (each expert in a different source task) to train a multi-task student network, by minimizing the cross-entropy loss between the student and experts' policies. To perform transfer, the resulting multi-task expert policy is used to initialize yet another DQN network, which learns the target task. The Actor-Mimic assumes that the source and target tasks share the same observation and action spaces, with different reward and transition functions. In \emph{Q-value reuse} \citep[Section 5.5]{Taylor2007a}, a Q-Learner uses $Q_{source}$ to kickstarts its learning of the target task, while also learning a new action-value function $Q_{target}$ to compensate $Q_{source}$'s irrelevant knowledge. In \citet{Taylor2007a}, the agent learns to play Keepaway games, and is introduced to a game with more players, resulting in more actions and state variables. \citet{Brys2015} transfers the source policy to a Q-Learning agent through reward shaping; the differences in the action and state spaces between source and target tasks are solved using a provided translation function.

Finally, Warm Start Reinforcement Learning (WSRL) \cite{Cheng2018,Zhu2017} aims at initializing the policy of the agent with another policy pre-trained on the same task. Domain knowledge, i.e., information about the environment known by the designers but not initially known by the agent, can be used to kickstart learning, either through imitation learning on expert demonstrations \cite{Cheng2018}, directly encoding it via propositional rules in the neural network architecture of the agent \cite{Silva2021}, or actively learning to imitate a transferred policy \cite{Wexler2022}.

Policy Intersection \cite{Griffith2013}, the method we evaluate in this paper, belongs to the category of Transfer Learning methods reviewed in the next section.

\subsection{Exploration}

Policy Shaping \cite{Griffith2013}, also referred to as guiding the exploration strategy of the agent, transfers a policy in a fast and effective way. Altering the exploration strategy is a popular technique in the safe RL domain, and consists in biasing or determining the actions taken by the agent at action selection time \cite{Garcia2015}. Such exploration requires the agent to be able to learn from \emph{off-policy} experiences, as it is the case with Soft Actor-Critic. Some existing work applies guided exploration to Transfer Learning \cite{Fernandez2006,Taylor2007b,Madden2004}, and illustrates how this technique allows the agent to outperform the performance of the source policy. Regarding the components of the source and target tasks that are allowed to differ, \citet{Fernandez2006} considers different goal placements (hence, different reward functions), \citet{Madden2004} uses symbolically learned knowledge to tackle states that are seen by the agent for the first time, and \citet{Taylor2007b} assumes similar state variables and actions, but a different reward function. In our setting, the environmental dynamics from the source task to the target task are completely disturbed and lead to wildly different outcomes following the actions of the agent, while the goal remains the same across tasks.

As our air compressor management problem can be translated into a Reinforcement Learning problem with continuous actions, in the next section, we propose a solution to extend Policy Intersection to continuous action spaces.

\section{Policy Intersection for Continuous Actions}

Equation \ref{eq:ps} expresses the product of two state-dependent probability distributions: the actor's $\pi_L(s_t)$, and the advisor's $\pi_A(s_t)$. In the discrete actions case, $\pi_L(s_t)$ and $\pi_A(s_t)$ are both vectors of size $|A|$, where $A$ is a finite set of actions available to the agent. Multiplying those two vectors amounts to taking the intersection between those two probability distributions. In other words, Policy Intersection samples actions that both the actor and the advisor ``agree on'', as these actions have a non-zero probability in both $\pi_L(s_t)$ and $\pi_A(s_t)$ \footnote{Note that if no agreement can be made, i.e., the intersection between which action the actor and the advisor want to choose is empty, it can be arbitrarily decided that either the advisor or the actor has the last word, depending on the problem.}. Unfortunately, when the action space is continuous, finding actions in the intersection between the actor's and the advisor's policies is less straightforward.

\begin{algorithm}[t]
        \caption{Policy Intersection for Continuous Actions}
        \label{alg:continuous-pi}
        \begin{algorithmic}
         \Require{$\pi_L$ is the currently learning actor, and $\pi_A$ is the frozen actor used as advisor}
         \State $A^L =$ set of actions sampled from $\pi_L(s_t)$
         \State $\forall a^L_i \in A^L:$ get probability density $\pi_A(a^L_i|s_t)$  \,
         \State Compute the Categorical Distribution $f(\pi_A(a^L_i|s_t))$ \,
         \If{$a^L_i \sim f(\pi_A) \neq None$}
            \State Execute action $a^L_i$ \,
        \EndIf
        \State Execute action $a \sim \pi_L(s_t)$
        \end{algorithmic}
\end{algorithm}

In Reinforcement Learning algorithms that can be applied to environments with continuous actions, such as SAC, individual actions can directly be sampled from the actor. Unfortunately, it is difficult to access a probability distribution over all possible actions, as the number of possible actions is infinite. However, the actor can provide, given a particular action, the probability density of that action. Assuming that the advisor can similarly be sampled, our approach to sampling the intersection between which actions are allowed by the actor and those allowed by the advisor is the following:

\begin{enumerate}
    \item If we use multiple advisors, one advisor is randomly picked from the set of advisors at each time-step.
    \item At action selection time, we sample a large amount of actions $A^L$ from the learned policy, or actor (where the superscript ``$L$'' stands for \emph{learned}).
    \item We then submit $A^L$ to the advisor $\pi_A$, which returns probability densities $\pi_A(s_t)$. For each action $a^L_i \in A^L$, we now have a corresponding density $\pi_A(a^L_i|s_t)$.
    \item We compute a Categorical distribution $f(\pi_A(a^L_i|s_t))$ over the discrete set of actions $A^L$, parameterized by the probability densities $\pi_A(a^L_i|s_t)$.
    \item We sample an action $a^L_i$ from $f(\pi_A(a^L_i|s_t))$. If no action can be sampled, then the policy of the actor $\pi_L$ is sampled.
\end{enumerate}

This method implements an ``AND'' between the actor and the advisor's policies. While our sampling technique allows to compute the intersection between an actor and an advisor, with no assumption about their distributions (we do not assume that they are Gaussian, for instance), we acknowledge that alternative approaches could have been devised, such as creating a new stochastic policy resulting of the point-wise multiplication of the advisor's probability density function with the actor's probability density function, sampled for a large number of actions.

\section{Reinforcement Learning Environment}
\label{sec:experiments_env}

A piston compressor is a type of industrial compressor, which increases the pressure of air enclosed in a cylinder. Air enters the cylinder through a valve, then is expelled under pressure by a piston operated by an eternal motor, through a second valve. In our particular setting, three compressors, each operated by its own motor, supply air to a tank. The air flows through a valve which is progressively opened or closed to simulate a demand. The demand curve changes at each episode. Episodes last for 250 time-steps; at each time-step, the agent chooses the rotary speed of each of the three motors operating the compressors. The action space is continuous, and ranges from 400 revolutions per minute (RPM) to 600 RPM, with the possibility to completely stop the motor if the agent chooses to. When restarted, the motor goes from 0 directly to 400 RPM. At each time-step, the agent receives an immediate reward consisting in the negative amount of energy (in joules) that was consumed to power the motors during the time-step. The agent must keep the air under a pressure in between 3 and 5 bars within the system; going above 5 bars does not bring any added value;  going under 3 bars triggers a backup policy. The backup policy is a naive controller that sets all three motors to their maximum speed for one time-step, which results in a very negative reward for the agent whenever the backup policy takes over. The agent observes the current speed of the motors, as well as the current pressure in bars, however, it does not know the demand curve. Although the reward is dense and highly informative, this setting consists in a challenging environment with continuous actions and partially observation states.

We generate several variants of this setting by randomly altering the size of the tank, as well as the compressors. The parameters determining the amount of air coming out of a compressor given $x$ RPM are defined at the beginning of an experiment, using the following formula: $\rho \times (1.0 + \lambda \times \zeta \sim \mathcal{R} \in [0, 1))$, where $\rho$ is a compressor parameter, $\lambda \in [0,1]$ is the percentage of alterations applied to $\rho$, and $\zeta$ is a random float yielded by a generator simulating a uniform distribution. The tank volume is defined as $50 + 10 \time \zeta$, in liters. The random float generator $\mathcal{R}$ takes an integer called \emph{seed} $s$ as parameter; we produce multiple variants of the environments to transfer across by varying the value of $s$ (from 1 to 17).

This simulated setup is modelled as an MDP as follows.

\subsection{Action space}

The action space is continuous and consists of 3 real values, ranging from -1 to 1. It is considered best practice [stable-baselines] to have the action and state spaces be centered around 0, and of a range of as close to 1 as possible. Each of the 3 real values allows to set the target RPM of its corresponding compressor, linearly interpolated between 0 (off, for an action value of -1) to 100\% (for an action value of 1).

The action space is discontinuous in two places: for every compressor, the actual effect of the action depends on the target RPM value it defines:

\begin{itemize}
    \item Below 20 RPM: the compressor is turned off completely
    \item Between 20 RPM and the minimum RPM of the compressor: the target RPM is adjusted to the minimum RPM of the compressor
    \item Above the minimum RPM of the compressor: the action is left untouched
\end{itemize}

The effect is that the agent can produce low actions to indicate that a compressor should work at its lowest RPM, and an even lower action to indicate that the compressor should be turned off. This allows the agent to control the on/off status of the compressors, in addition to their target speed, with a single real value.

\subsection{Observation space}

The observation space consists of several real values that measure past tank pressures and RPMs. For every time-step, 4 real values are logged: the current pressure (in bars) in the tank, and the current RPMs of the 3 compressors. When producing observations, the environment looks $N$ time-steps in the past to produce $N$ real values corresponding to the tank pressures at these past $N$ time-steps (so, this is a history of past tank pressures). The environment also looks $M$ time-steps in the past to produce $3M$ real values corresponding to RPMs of the compressors during these past $M$ time-steps. $N$ and $M$ can be distinct, and in our experiment, we use $N = 5$ past tank pressures and $M = 3$ past RPMs. %% TODO: Update with actual experiments.

By observing past tank pressures and RPMs, the agent gets a feel of:

\begin{itemize}
    \item How fast is the tank depleting, which allows it to approximate the derivative of the tank pressure (that only looks at the past), thereby adjusting the future RPMs.
    \item What is the demand of air in the recent past (by combining how the tank pressure changes and what the RPMs were), which may help the agent guess what the demand will be in the future if there are time-specific patterns in the demand.
\end{itemize}

This information is still not enough for optimal control, as it does not contain information about the future demand, but observing a history of past sensor readings has been shown to be one of the best approaches to learn in Partially Observable MDPs [citation], and is easy to implement.

\subsection{Reward function}

The reward function is the change in cost that occurs after a given time-step executes (so, after 60 seconds of simulated time after an action has been applied to the physical or simulated system). More specifically, the reward given after a time-step consists of two components:

\begin{enumerate}
    \item Minus the amount of kilojoules consumed during the time-step. In the simulated setup, the power consumption of the compressors is approximated in a lookup table (in watts). In simulation, we assume that a compressor instantly reaches its target RPM and its corresponding power consumption, so the amount of kilojoules is $60 s \times P \times 0.001$, with P the power (in watts) obtained from the lookup table.
    \item A turn on penalty when a compressor turns on. In addition to the point above (the power consumption during the whole time-step), a compressor that goes from off to on at the start of a time-step is considered to incur a cost equal to its maximum (max RPM) power consumption during 60 seconds. For instance, if the third compressor powers on, a reward of $-60 s \times 2000 W \times 0.001 = -120 kJ$ is given to the agent.
\end{enumerate}

\section{Experiments}
\label{sec:experiments_results}

% TODO explain how the plots are computed
% LUT 5 has been advised by all the other seeds
% All 100 demand curves are used (no train/test split)
%
% MPC baseline
%
% MPC on seed 5 with model also using seed 5 (perfect model): cost of 635 on average for all demand curves
% MPC on seed 5 with model using other seeds: cost of 753
% Average over all seeds of MPC on its own seed: cost of 763
% Average over all seeds of MPC with the model being another seed: cost of 1050

We set $\lambda$, the percentage of alteration applied to the parameters of the compressors, to 0.5. We train 16 different advisory policies, each trained on the environment with a different seed $s \in [1, ...16]$. We consider two settings in which Policy Intersection for continuous actions can be used: i) (\emph{single}) a single advisor, with a different seed than the advisee, is exploited, ii) (\emph{multiple}) all available advisors are exploited, except the one with the same seed as the advisee. To leverage multiple advisors, at each time-step, one advisor is randomly picked from the set of advisors. We compare transferring advisory policies using Policy Intersection to simply loading an advisor of a different seed and letting it resume learning.

\begin{figure}[!t]
\centering
\includegraphics[width=3.in]{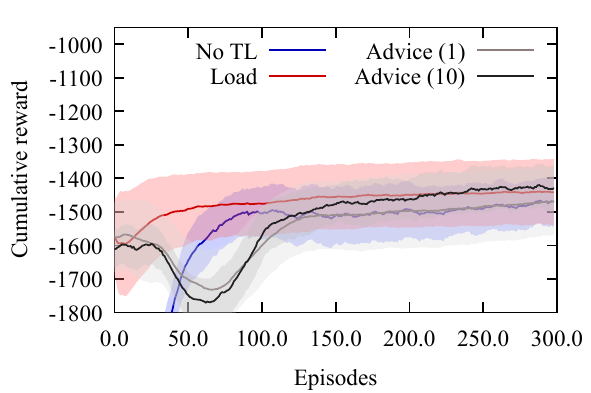}
\caption{Comparison between transferring from 16 advisors, 1 advisor, loading one advisor at a time, or not using any transfer at all. Using Policy Intersection, either with a single or multiple advisors, outperforms loading an advisor and letting it resume learning. However, there is a significant dip in the learning curve shortly after learning begins when Policy Intersection is used.}
\label{fig_loadvsadvised}
\end{figure}

\begin{figure}[!t]
\centering
\includegraphics[width=3.in]{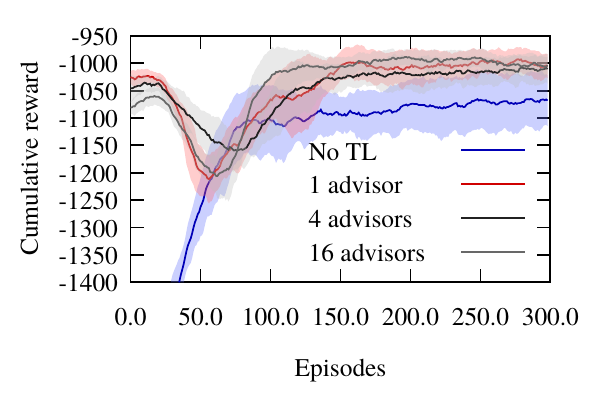}
\caption{There is a significant difference in performance between using 4 and 16 advisors, especially when recovering from the dip in the curve. This difference lessens in the long run.}
\label{fig_advisedonevsmany}
\end{figure}

At first glance, in both Figure \ref{fig_loadvsadvised} and Figure \ref{fig_advisedonevsmany}, a significant dip can be noticed in the learning curve, occurring only a few episodes after the beginning of learning. Moreover, this dip only occurs when using Policy Intersection; it is not present when loading an advisor. This decrease in performance is a property of off-policy algorithms, such as Soft Actor-Critic, when their learning is kickstarted by an already pre-trained policy, as in this experiment. At the beginning of learning, the policy of the advisee $\pi_L$ is random, while the advisor $\pi_A$ is not, and shows actions to the advisee that are generating good rewards. At this point, the critic of Soft Actor-Critic believe that all actions are uniformly good, since the advisee has only seen good actions, thanks to the advisor's guidance. Due to the inner randomness of neural networks, the policy of the advisee then arbitrarily concentrates its probability on one action, which is not necessarily good, producing the dip in the curve. It takes some exploration time for the advisee to realize the suboptimality of that action, and to learn to choose better ones, hence the recovering occurring a few episodes later. We do not propose a technique to mitigate this particular problem in this paper, but solutions have been investigated \cite{Wexler2022}.

Policy Intersection, leveraging either a single or multiple advisors, outperforms loading an advisor and letting it resume learning. In addition, we observe in Figure \ref{fig_loadvsadvised} that exploiting multiple advisors instead of a single one significantly improve performance, especially when recovering from the dip in performance. This phenomenon is also noticeable in Figure \ref{fig_advisedonevsmany}, in which we compare leveraging 1 advisor, 4 advisors, and 16 advisors. However, there is a performance ceiling, reached by all settings in the long run, that does not seem possible to exceed by increasing the amount of advisors.

\section{Conclusion}

We applied Soft Actor-Critic to learn to control three motors, each activating a piston compressor in an air compressor management problem. In this problem, air is expelled by the piston compressors in a tank, while air is going out through a valve which progressively opens or closes following a demand curve. The goal of the compressors is to provide a satisfactory pressure in between 3 and 5 bars, while conserving energy. Moreover, we considered the problem of learning a new controller based on previously learned ones on different variants of the air compressor management problem. To tackle this Transfer Reinforcement Learning problem, we used Policy Intersection, a Policy Shaping method, allowing one or multiple already acquired policies to distill their knowledge in a new policy, through guiding the exploration strategy of the agent. We empirically showed that this approach outperforms simply loading a transferred controller, and that exploiting multiple transferred controllers instead of a single one can significantly improve performance.
In future work, we will focus on finding and comparing solutions to mitigate the degradation of performance occurring a few episodes after the beginning of learning, a phenomenon known as \emph{Warm Start Reinforcement Learning Degradation} in \cite{Wexler2022}.
\section*{Acknowledgment}

The first author is funded by the Science Foundation of Flanders (FWO, Belgium) as 1SA6619N Applied Researcher.

\vfill
\pagebreak

\bibliography{main}
\bibliographystyle{tmlr}

\end{document}